\documentclass{sig-alternate-05-2015}
\usepackage{latexsym}
\usepackage{amsmath,bm}
\usepackage{amsfonts}
\usepackage{mathrsfs}
\usepackage{amsbsy,amssymb}

\usepackage{graphicx}
\usepackage{xcolor}
\usepackage{multirow}
\usepackage{paralist}
\usepackage{subfigure}
\DeclareMathOperator{\softmax}{softmax}
\DeclareMathOperator{\vect}{vec}
\DeclareMathOperator*{\minimize}{minimize}
\DeclareMathOperator{\transpose}{^{\!\!\top}}
\newcommand{\tabincell}[2]{\begin{tabular}{@{}#1@{}}#2\end{tabular}}

\begin{document}

\setcopyright{acmcopyright}

\doi{10.475/123_4}




%

\title{Distilling Word Embeddings: An Encoding Approach}

\numberofauthors{6} 
%
\author{
Lili Mou, Ran Jia, Yan Xu, Ge Li,$^*$ Lu Zhang, Zhi Jin$^*$\\
\affaddr{Key Laboratory of High Condence Software Technologies (Peking University), MoE, China}\\
\affaddr{Institute of Software, Peking University, China\quad $^*$Corresponding authors}\\
\email{\{doublepower.mou, jiaran1994\}@gmail.com}\\
\email{\{xuyan14, lige, zhanglu, zhijin\}@sei.pku.edu.cn}
}

\maketitle

\begin{abstract}
Distilling knowledge from a well-trained cumbersome network to a small one has recently become a new research topic, as lightweight neural networks with high performance are particularly in need in various resource-restricted systems. This paper addresses the problem of distilling word embeddings for NLP tasks. We propose an encoding approach to distill task-specific knowledge from a set of high-dimensional embeddings, which can reduce model complexity by a large margin as well as retain high accuracy, showing a good compromise between efficiency and performance. Experiments in two tasks reveal the phenomenon that distilling knowledge from cumbersome embeddings is better than directly training neural networks with small embeddings.
\end{abstract}

%
%


%
%

%
%



\section{Introduction}

Distilling knowledge from a neural network---that is, 
transferring valuable knowledge from a cumbersome network to a 
lightweight one---is pioneered by Bucilu\v{a} et al.~\cite{modelcompression};
it has attracted increasing attention 
over the last two years \cite{distilling,hashnet}.

As addressed by Hinton et al.~\cite{distilling}, 
the objective of training networks is probably 
different from deploying networks: during training
we focus on extracting as much knowledge as possible from a large dataset,
whereas deploying networks takes into consideration multiple aspects,
including accuracy, memory, time, and energy consumption.
It would be appealing if we can 
first well train a cumbersome network offline, and then distill its knowledge to a small one
for deployment.
The aim of knowledge distillation is thus to reduce model complexity as well as to 
retain high performance,
which is particularly important to neural networks' applications
in resource-restricted scenarios, e.g., real-time systems, mobile devices, and
large ensembles of models.

\begin{figure*}[!t]
\centering
\includegraphics[width=.8\textwidth]{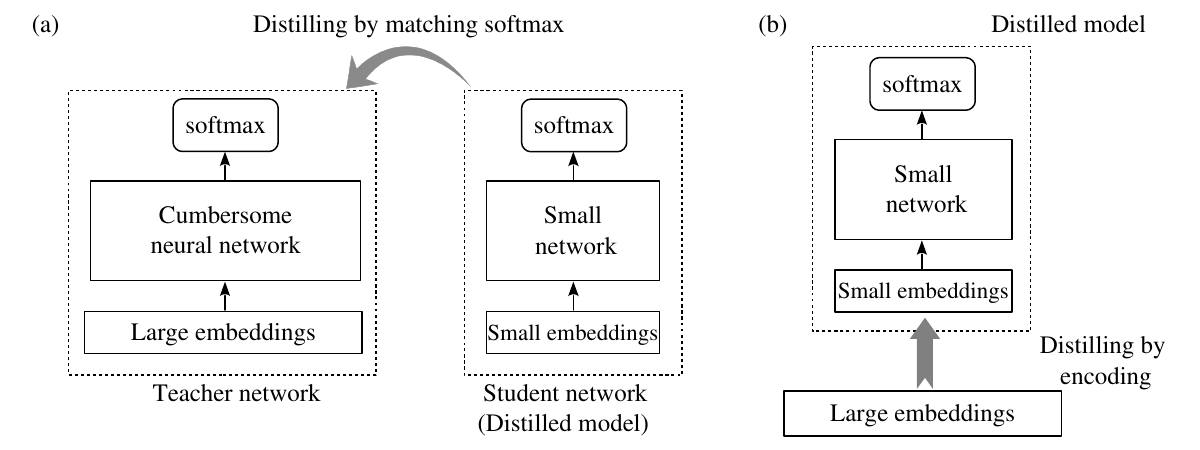}
\caption{Distilling knowledge by (a) matching softmax, and (b) encoding embeddings.}
\label{fDistill}

\end{figure*}

Much evidence in the literature shows the feasibility of transferring
knowledge from one neural network to another, for instance, from shallow networks to deep ones \cite{fitnet},
from feed-forward networks to recurrent ones \cite{dnn2rnn}, 
or vice versa~\cite{deep2shallow,rnn2dnn}.
The main idea of the above studies is to train a \textit{teacher model} first, and then use the 
teacher model's output
(estimated probabilities by $\softmax$, say, in a classification problem) 
to guide a \textit{student model}. 
Several variants of training objectives include applying regression over the input of $\softmax$ \cite{deep2shallow} and
softening the teacher model's probabilities \cite{distilling}.
We call such approaches ``matching $\softmax$'' (Figure~\ref{fDistill}a).
It is also argued that 
the estimated probabilities by a teacher model
convey more information than one-hot represented ground truth; hence
knowledge distillation is feasible and beneficial~\cite{distilling}.

Despite the above generic approach,
this paper focuses on distilling word embeddings in NLP applications.
Particularly, we find the specificity of embeddings brings new opportunities
for knowledge distillation.

As word embeddings map discrete words to distributed, real-valued vectors, it can be viewed that a word
is first represented as a one-hot vector and then the vector 
is multiplied by a large embedding matrix, known as a look-up table.
During the matrix-vector multiplication,
one and only one column in the look-up table is verbatim retrieved 
for a particular word.
Thus, we may build an interlayer---sandwiched between
the high-dimensional embeddings and the ensuing network---to squash embeddings to a low-dimensional space
(Figure~\ref{fDistill}b). 
The standard cross-entropy loss can then be applied to train the encoding layer
and other parameters in the network.
In such a supervised manner, task-specific knowledge in the original cumbersome embeddings
can be distilled to low-dimensional ones.

In summary, the main contributions of this paper are three-fold: (1)
We address the problem of distilling word embeddings in NLP applications.
(2) We propose a supervised encoding approach to distill
task-specific knowledge from cumbersome word embeddings.
(3) Our experimental results in sentiment analysis and relation classification tasks reveal a phenomenon that 
distilling low-dimensional embeddings from large ones
is better than directly training a network with small embeddings.

It should also be noticed that the proposed encoding approach does not rely on a teacher model;
our method is complementary to existing matching $\softmax$ for knowledge distillation.

\section{Background of Matching Softmax}\label{sBackground}

As said, existing approaches to knowledge transfer between two neural networks
mainly follow a two-step strategy: first training a teacher network; then using
the teacher network to guide a student model by matching $\softmax$,
depicted in Figure \ref{fDistill}a.

For a classification problem, $\softmax$ is typically used as the output
layer's activation function. Let $\bm z\in\mathbb{R}^{n_c}$ be the input of $\softmax$.
($n_c$ is the number of classes.)
The output of $\softmax$ is
$$y_i=\dfrac{e^{z_i/T}}{\sum_{j=1}^{n_c}e^{z_j/T}}$$

\noindent where $T$ is a relaxation variable (used later), called \textit{temperature}.
$T=1$ for standard $\softmax$.

Take a 3-way classification problem as an example. If a teacher model estimates
$\bm y=(0.95, 0.04, 0.01)\transpose$ for three classes, it is valuable information to the student model that 
Class 2 is more similar to Class 1 than Class 3 to Class 1.

However, directly imposing constraints on the output of $\softmax$ may be ineffective:
the difference between 0.04 and 0.01 is too small.
Ba et al.~\cite{deep2shallow} match the input of $\softmax$, $\bm z$, rather than $\bm y$.
Hinton et al.~\cite{distilling} raise the temperature $T$ during training, which makes the estimated probabilities
softer over different classes. The temperature of 3, for instance, softens
the above $\bm y$ to $(0.64, 0.22, 0.14)\transpose$.
Matching $\softmax$ can also be applied along with standard cross-entropy loss
(with one-hot ground truth), or more elaborately,  
the teacher model's effect declines in an annealing fashion
when the student model is more aware of data \cite{fitnet}.

\section{The Proposed Encoding Approach for Distilling Embeddings}\label{sApproach}

This section introduces in detail our proposed method for word embedding distillation
(Figure \ref{fDistill}b). We also analyze the neural network's model capacity with distilled embeddings, and discuss the rationale for 
distilling small embeddings from large ones in a supervised manner, instead of directly training with small embeddings.

Word embeddings are a standard component for neural natural language processing. As feeding word indexes directly to neural networks is somewhat nonsensical, words are mapped to a real-valued vector,
called \textit{embeddings}, where each dimension captures a certain aspect of underlying word semantics. Usually, they are trained in an unsupervised fashion, e.g., maximizing the probability of a large corpus \cite{LM,hierarchical}, or maximizing a scoring function \cite{unified,word2vec}. 
The learned embeddings can be fed to standard neural networks for supervised learning,
e.g., POS tagging, named entity recognition, and semantic role labeling \cite{unified}.

To formalize word embeddings in algebraic notations,
we let $\bm x_i\in\mathbb{R}^{|\mathscr{V}|}$ be one-hot representation of the $i$-th word $x_i$
in the vocabulary $\mathscr{V}$; the $i$-th element in the vector $\bm x_i$ is on, with other
elements being 0.
Let $\Phi_c\in\mathbb{R}^{n_{\text{embed}}\times |\mathscr{V}|}$ be a (cumbersome) embedding matrix (look-up table).
Then the vector representation of the word is exactly the $i$-th column of the matrix, given by 
$\Phi_c\cdot\bm x_i$.

Now we consider distilling, from cumbersome embeddings $\Phi_c\cdot\bm x_i$, an $n_\text{distill}$-dimensional vector for the word, where $n_\text{distill}$ is smaller than $n_\text{embed}$. 
It is accomplished by encoding with a non-linear neural layer, i.e.,
\begin{equation}
\vect(x_i)=f(W_{\text{encode}}\cdot\Phi_c\bm x_i+\bm b_{\text{encode}})\label{eEncode}
\end{equation}
where $W_{\text{encode}}\!\in\!\mathbb{R}^{n_\text{distill}\times n_\text{embed}}$
and $\bm b_{\text{encode}}\!\in\!\mathbb{R}^{n_\text{distill}}$
are parameters of the encoding layer; $\vect(\cdot)$ denotes the distilled
vector representation of a word.

These distilled embeddings can then 
be fed to a neural network (with parameters $\Theta$) for further processing.
Let $m$ be the number of data samples and $n_c$ be the number of target classes;
suppose further $\bm y^{(j)}$ is the output of $\softmax$ 
for the $j$-th data sample and $\bm t^{(j)}$ the one-hot represented ground truth.
Our training objective is the standard cross-entropy loss, given by
$$\minimize\limits_{W_{\text{encode}},\bm b_{\text{encode}},\Theta,\Phi}
\ -\sum_{j=1}^m\sum_{i=1}^{n_c}t_i^{(j)}\log y_i^{(j)}$$

We would like to point out that distilling embeddings does not
increase, or in fact may reduce, model capacity vis-\`a-vis directly training
with small embeddings, despite the large number of parameters
in cumbersome embeddings and the coding layer's weights.

\medskip
\noindent\textbf{Theorem 1.} \textit{
The model capacity of a neural network with distilled embeddings is less than or equal to
that of a neural network trained directly with small embeddings.}

\smallskip
\noindent\textit{Proof.} 
The intuition is straightforward: small embeddings are free parameters which are not constrained,
whereas the distilled embeddings are subject to the form in Equation~\ref{eEncode}.
Formally, let $\mathcal{H}_d, \mathcal{H}_s$ be the hypothesis classes of networks with 
distilled$/$small embeddings,
respectively. For each $h_d\in\mathcal{H}_d$ with cumbersome embeddings 
$\Phi_c$ and encoding parameters $W_{\text{encode}},\bm b_{\text{encode}}$, 
there exists a hypothesis $h_s\in\mathcal{H}_s$, satisfying that $h_s=h_d$ with small embeddings $\Phi_s$, whose $i^{\text{th}}$ column (the small embedding for $i^{\text{th}}$ word) is $f(W_{\text{encode}}\Phi_c\bm x_i$ $+\bm b_{\text{encode}})$. Hence, $\mathcal{H}_d\subseteq\mathcal{H}_s$.\quad \rotatebox{90}{$\blacktriangle$}

\bigskip
A curious question is then why distilling embeddings may help,
compared with directly training the neural network with small embeddings.
We provide an intuitive explanation as follows.

Since word embeddings are typically learned from a large corpus in an unsupervised manner,
the knowledge in embeddings is restrained by dimensionality.
For example, the sentiment of a word is of secondary importance compared with
its syntactic functionality in a sentence.
Hence, sentiment information might be lost
in low-dimensional embeddings, which is unfavorable in a sentiment analysis task.
On the contrary, large embeddings have the capacity to 
capture different aspects of word semantics.
The proposed supervised encoding approach may then distill task-specific (e.g., sentiment)
knowledge to a small space, while eliminating irrelevant information.
Therefore, we may reasonably expect that 
distilling embeddings would outperform direct use of small ones.

\subsubsection*{Deployment Issues}

Before deploying the model, we shall precompute the distilled embeddings, $\vect(\cdot)$, according to Equation~\ref{eEncode} after training all parameters. The original embeddings $\Phi_c\bm x$ and encoding parameters ($W_\text{encode}$ and $b_\text{encode}$) can then be safely discarded, and we obtain a small model (dashed rectangle in Figure~\ref{fDistill}b) with a set of small embeddings, which are distilled from large ones.

As we shall see in the experiments, the small model will be very computational efficient because we have reduced a large number of parameters.

\section{Evaluation}

In this section, we present our experimental results.
We first describe the testbed and protocol of our experiments in Subsection \ref{ssTestbed}.
Then we analyze in Subsection \ref{ssResult} the performance of our approach 
regarding several aspects, namely accuracy, memory, and time consumption.


\subsection{Tasks, Models, and Protocols}\label{ssTestbed}

We tested our distilling approach in two tasks:
sentiment analysis and relation classification.

The sentiment analysis task aims to classify a sentence into 5 categories according to its
sentiment: strongly/weakly positive/negative and neural.
We used Stanford Sentiment Treebank\footnote{http:/\!/nlp.stanford.edu/sentiment/} 
as our dataset, which contains 8544/1101/221 sentences for training, validation, and testing.
Phrases (sub-sentences) in the training set are also labeled with sentiment, 
enriching the training set to more than 150k samples.
For validation and testing, only the sentiment of a whole sentence was considered.

The second task is to classify the relation between two tagged entities in a sentence.
The SemEval 2010 dataset,\footnote{http:/\!/semeval2.fbk.eu/semeval2.php?location=data} we used, comprises 8000 training samples, from which we split
10\% for validation; there are additional 3000 samples for testing.
Target labels include 9 directed relations (e.g. {\tt Component-Whole}) plus a default {\tt Other}; in total, we have 19 classes.
The official $F_1$-score was applied as our measurement.

To set up our experiments, we leveraged two state-of-the-art neural models: a tree-based convolutional neural network (TBCNN) for sentiment 
analysis \cite{tbcnn_sentence}, and a long short term memory-based recurrent network along 
shortest dependency path (SDP-LSTM)\footnote{
We only used
word embeddings, and ignored other features like 
hyponymy, dependency types, which were used in \cite{sdp-lstm}.
In this way, we focus on the problem of embedding distillation itself.}
between two entities for relation classification \cite{sdp-lstm}.

For each task, we evaluated our proposed methods by distilling 300-dimensional 
embeddings to 50 dimensions, further processed by a thin network (also 50d).
In comparison, we trained the 50d network directly with small 50d embeddings.
All models were trained by mini-batch gradient descent with back-propagation.
For both settings of distilling and non-distilling, we tried extensive 
configurations of hyperparameters, mainly following the original papers.\footnote{
Due to the limitation of space, we list candidate configurations on our website:\\
https:/\!/sites.google.com/site/distillembeddings/
} After choosing the setting with the highest validation accuracy, we ran
each model 5 times for smoothing with different random initializations, and report the average test accuracy or $F_1$-score.

\subsection{Results}\label{ssResult}
\begin{table}[!t]
\centering
\resizebox{.476\textwidth}{!}{
\begin{tabular}{|l|l||c|c|c|}
\hline
\textbf{Task}     & \textbf{Method} & \textbf{Acc.}  & \textbf{$\#$Param} &\textbf{Time}\\
\hline
\hline
\multirow{4}{*}{\tabincell{l}{Sentiment\\ analysis\\by TBCNN}} 
                          & Cumbersome embed. &  51.6 &   6.9M  & 1x \\
\cline{2-5}
                         & Small embed.      &  46.4   &\multirow{3}{*}{\tabincell{c}{0.94M\\(0.14x)}} & \multirow{3}{*}{0.04x}\\
                         & Distilled embed.  & 47.5          &  &\\
                         & Matching $\softmax$  & 45.8  & &\\
\hline
\end{tabular}
}
\caption{Comparison between cumbersome embeddings, small embeddings, matching softmax, and distilled embeddings. The official measure is accuracy (acc.) in percentage.\label{tCompare0}
}
\end{table}

\begin{table}[!t]
\centering
\resizebox{.476\textwidth}{!}{
\begin{tabular}{|l|l||c|c|c|}
\hline
\textbf{Task}     & \textbf{Method} &  \textbf{$F_1$}  & \textbf{$\#$Param} &\textbf{Time}\\
\hline
\hline
\multirow{5}{*}{\tabincell{l}{Relation\\ classification\\by SDP-LSTM}} 
               & Cumbersome embed. & 82.1 &  8.8M & 1x \\
\cline{2-5}
              & Small embed.      &  79.0 & \multirow{4}{*}{\tabincell{c}{1.3M\\(0.15x)}} & \multirow{4}{*}{0.04x}  \\
             & Distilled embed.  &  79.4 &              &   \\
             & Matching $\softmax$ &  80.1 & & \\
             & Hybrid & 80.2 & &\\
\hline
\end{tabular}
}
\caption{Comparison between cumbersome embeddings, small embeddings, matching softmax, and distilled embeddings. We further made an attempt to combine matching softmax and distilling embeddings (denoted as ``Hybrid''). The official measure for relation classification is the $F_1$-score.
}
\label{tCompare}
\end{table}

Tables \ref{tCompare0} and \ref{tCompare} presents the results of our proposed model as well as
two competing settings: training a wider network with cumbersome embeddings,
and directly training a thin network with small embeddings.

In both experiments, cumbersome embeddings yield the highest performance,
the distilled embeddings rank second, and small embeddings are worst.
Basically, our method outperforms direct training of a small network
by a margin of approximately one standard deviation (std $=$ 1.1 and 0.6, respectively).
As the results were obtained by averaging over 5 initializations, we deem the improvement is fair.

Regarding model complexity, our distilled embeddings 
reduce memory and time to a large extent to 14--15\% and 4\%, respectively
({\tt C++} implementation on a single CPU).
Therefore, the resulting network is significantly more lightweight,
which is helpful to deployment in neural networks' applications.

\begin{figure}[!t]
\centering
\includegraphics[width=.34\textwidth]{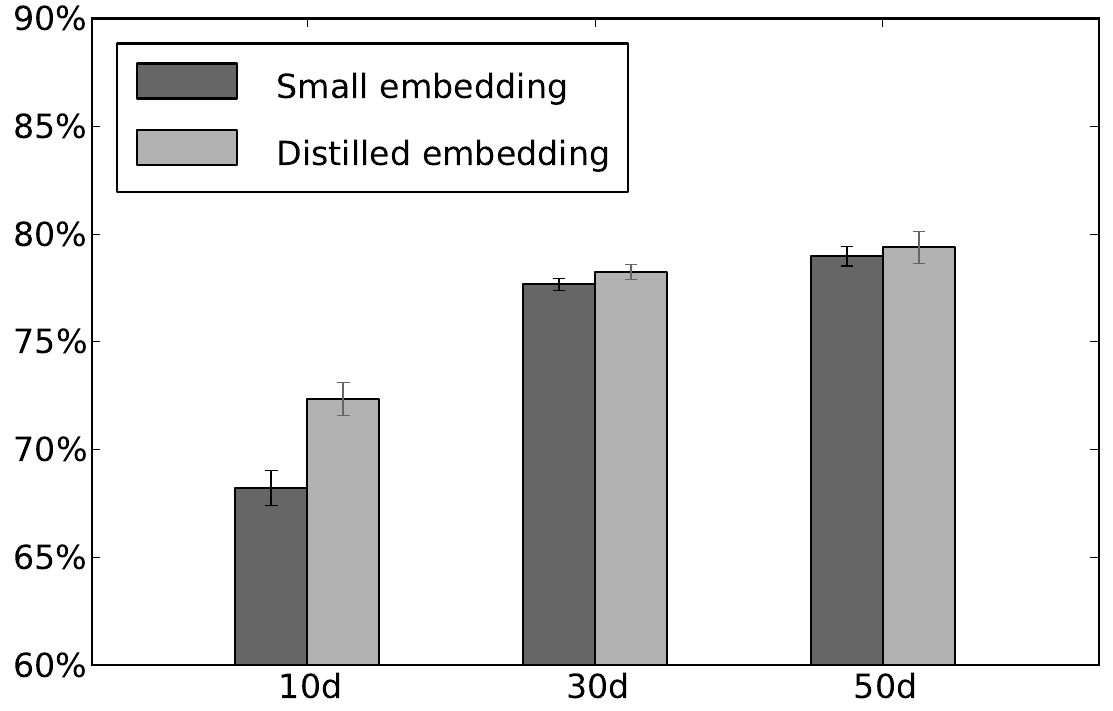}
\vspace{-.4cm}
\caption{Accuracy versus dimension in the experiment of relation classification.}\label{fCurve}
\vspace{-.3cm}
\end{figure}

To further test our method under extreme conditions,
we distilled word embeddings to 10d and 30d. We chose to conduct the experiments in the second task, because it is of lower variance.
As demonstrated in Figure~\ref{fCurve}, our method
consistently outperforms direct training with small embeddings in all scenarios; moreover,
the margin increases when the dimension becomes small.
Such result is consistent with our human intuition, and verifies the conjecture
in Section \ref{sApproach}---small embeddings contain less knowledge specific to the
task of interest; the proposed supervised encoding approach can distill task-specific knowledge from
large embeddings.

We also notice that our approach is, in fact, complementary to existing matching $\softmax$ methods: the encoding layer
distills task-specific knowledge from large embeddings in a bottom-up fashion, 
whereas matching $\softmax$ distills generic knowledge in a top-down fashion.

In both tasks, we also tried the matching $\softmax$ approach, whose settings and hyperparameters are mainly derived from~\cite{distilling}, i.e., $T=2$ and a $1\!:\!1$ mixture of ground truth and the teacher model's output. Its performance is not consistent: in the sentiment analysis task, matching $\softmax$ hurts the performance by 0.6\%, whereas it improves the relation classification task by 1.1\%. (See Tables~\ref{tCompare0} and \ref{tCompare} again.) One plausible explanation is that the teacher model itself has not achieved remarkable accuracy (only about 50\%) in the 5-way sentiment classification task. Using a teacher model introduces additional knowledge as well as errors. If the latter dominates, matching $\softmax$ may hurt the student model.
However, our encoding approach to embedding distillation does not reply on a teacher model. Combined with matching $\softmax$, it improves another 0.1\% (although may not be large) in the second experiment, showing that the two methods can be potentially combined, as they are complementary to each other.

\section{Conclusion}

In this paper, we addressed the problem of distilling embeddings for NLP,
which is important when deploying a neural network in resource-restricted scenarios.
We proposed an encoding approach that distills cumbersome word embeddings to
a low dimensional space. Experimental results have shown the superiority of our 
proposed distilling method to training neural networks directly with small embeddings; that the performance gain increases significantly especially when the dimension becomes small.
Moreover, our approach does not reply on a teacher model, which is complementary to matching softmax; these two methods of knowledge distillation could also be combined.

\section{Acknowledgments}

We thank all reviewers for their insightful comments and Rui Yan for discussion of the manuscript.
This research is supported by the National Basic Research Program of China (the 973 Program) under Grant No.~2015CB352201, the National Natural Science Foundation of China under Grant Nos.~61232015, 91318301, 61421091, and 61502014.




\bibliographystyle{abbrv}
\bibliography{distill}
\end{document}